\documentclass[10pt,twocolumn,letterpaper]{article}

\usepackage{3dv}
\usepackage{times}
\usepackage{epsfig}
\usepackage{graphicx}
\usepackage{amsmath}
\usepackage{amssymb}

\usepackage[pagebackref=true,breaklinks=true,letterpaper=true,colorlinks,bookmarks=false]{hyperref}

\threedvfinalcopy 

\def\threedvPaperID{80} 

\usepackage[utf8]{inputenc} 
\usepackage[T1]{fontenc}    
\usepackage{hyperref}       
\usepackage{url}            
\usepackage{booktabs}       
\usepackage{amsfonts}       
\usepackage{nicefrac}       
\usepackage{microtype}      

\usepackage{enumitem,amssymb}
\newlist{todolist}{itemize}{2}
\setlist[todolist]{label=$\square$}
\usepackage{soul} 

\usepackage{subcaption}
\usepackage{wrapfig}
\graphicspath{ {./images/} }

\begin{document}
\title{ Video based real-time positional tracker }

%

\author{%
  David Albarrac\'in\\
{\tt\small dalbarracin@virtuallylive.com}
\and
  Jes\'us Hormigo\\
{\tt\small jesus@virtuallylive.com}

\and
  Jos\'e David Fern\'andez \\
{\tt\small jdfernandez@virtuallylive.com}
}

\maketitle

\begin{abstract}
We propose a system that uses video as the input to track the position of objects relative to their surrounding environment in real-time. The neural network employed is trained on a 100\% synthetic dataset coming from our own automated generator. The positional tracker relies on a range of 1 to \textit{n} video cameras placed around an arena of choice.

The system returns the positions of the tracked objects relative to the broader world by understanding the overlapping matrices formed by the cameras and therefore these can be extrapolated into real world coordinates.

In most cases, we achieve a higher update rate and positioning precision than any of the existing GPS-based systems, in particular for indoor objects or those occluded from clear sky.
\end{abstract}

\section{Introduction}
\label{section_introduction}

In recent years, location technology has advanced dramatically with electronic devices becoming more robust, compact and portable. The miniaturization of circuitry and effectiveness of positional tracking systems and the new types of self-contained instruments can now track position, time and direction of motion more effectively. In addition to GPS, the self-calibrating gyroscopes, accelerometers, inertial motion units (IMU) and other methods \cite{koyuncu2010survey} can help with occlusion and poor signal problems. Radio-frequency contamination can be a problem to obtain the signal from the object that’s being tracked as well.

The low frequency on position updating that the traditional GPS chips can produce \cite{salih2013suitability}, implies that tracking fast moving objects is a mission impossible in many cases. In sports where fast paced movements happen, but also when they take place in a sky-occluded arena or indoors, satellite technology is not be reliable or available. Higher precision location devices, such as RTK GPS \cite{langley1998rtk}, exist in the market but its strong distance-dependence, setup costs and the need of calibration against a base station, make it a complex and expensive solution to implement for the purpose. On the other hand, visual tracking has been a challenging problem in computer vision over the years \cite{smeulders2013visual, yilmaz2006object}.

In this paper we describe our proposal for a self-contained system that tracks multiple objects and people in one or multiple video feeds.

\section{Related work}
\label{section_relwork}

Positional tracking has been tackled from multiple approaches based on computer vision, such as feature based and more recently deep learning \cite{smeulders2013visual}.

With the evolution of deep learning, scientist and researchers have developed different positioning systems that learn to directly regress the absolute camera pose in the scene from a single or from multiple images \cite{scenecoord}. \cite{contextualnet} developed a recurrent neural network based system that learns the spatial and temporal features and estimates the precise camera pose from a sequence of images.

Most of the positioning systems learn from the images captured in the past and hence are not able to accommodate for environmental changes that are dynamic in nature. The main contributions of our work can be summarized as follows:

\begin{itemize}
\itemsep0em 
    \item Autonomously trained system without requiring the labor intensive process of labeling data for training the system as the image captured.
    \item Real-Time tracking system ($\sim$24Hz) - Can be increased with more powerful hardware.
    \item High accuracy positioning with a max deviation of 25cm  - Proportional to resolution and number of cameras.
    \item Indoor/Outdoor usability without clear-sky conditions or radio-frequency artifacts.
\end{itemize}

\section{Positional tracker}
\label{section_postrack}

In this section we walk through the challenge of positioning objects within an absolute coordinate system, exploiting CV in parallel. We broke down the problem in three incremental steps. 

First, we adapted a traditional object detection NN to provide, instead of just 4-coordinates bounding boxes relative to the plane of the camera, additional points that are useful to describe the object(s) within the real environment as well. Once we achieved inference of the 3D positional information of the object within the camera's default coordinate system, we then inferred the object orientation to the camera's coordinate system. To achieve this, we required further study and restructured the anchors and prediction tensor mechanisms, given the higher complexity and non-euclidean nature of $SO(3)$. We then saw how these were fitted in the larger coordinate system of the real world by looking at the overlapped regions.

\subsection{Background}

We had developed a system that is regularly used on live TV and that automates object tracking and info-graphic generation in real-time by using neural networks. One of the main challenges we faced during its development was the network training, given that there there are no large (or small) dataset for the objects we needed to track. As a solution, we designed a synthetic images generation tool, capable of tagging and produce datasets at a massive scale. We have found that systems trained on a mix of real and synthetic images are able to detect the objects in a more effective way than using only real or synthetic. A second problem we solved was the optimization of the inference pipeline, given that  executing this as fast as possible enabled several ways of improving the system and its final outcome.

Under these conditions, the next goal was to infer more information about the object appearing on-screen; in particular, inferring 3D information would enable new and exciting possibilities ranging from infographic overlays pointing to specific components of the objects to varied special effects with precision to the object, such as adding smoke or fire, for example.
As an alternate way of taking advantage of this 3D information, we forked the project into a new one, with the goal of completely describing in 3D and geolocate objects inside an arena with coverage of one or more regular video cameras. While there have been many attempts to achieve a similar goal, we used an original system for real-time object tracking as a basis, which was already on the market, ensuring that our solution struck a sweet spot between inference speed and accuracy unlike any other system.

\subsection{Synthetic dataset Generation (SynGen)}
Datasets for specific cases remains scarce \cite{dubuisson2016survey}, requiring these to be created from scratch. This represents a bottleneck on the process of training a custom neural network. An advantage of the automatic synthetic dataset generation is that, with relatively little effort \cite{sun2015generating}, labeling can cover different disciplines of tagging valid for multiple problems, such as classification, detection or segmentation, with the most precision possible while still being able to add random noise in controllable conditions. From a programming point-of-view, versatility and address-ability make this essential for a precise training of the networks.

\begin{figure}[t]\begin{center}
\includegraphics[width=1\linewidth]{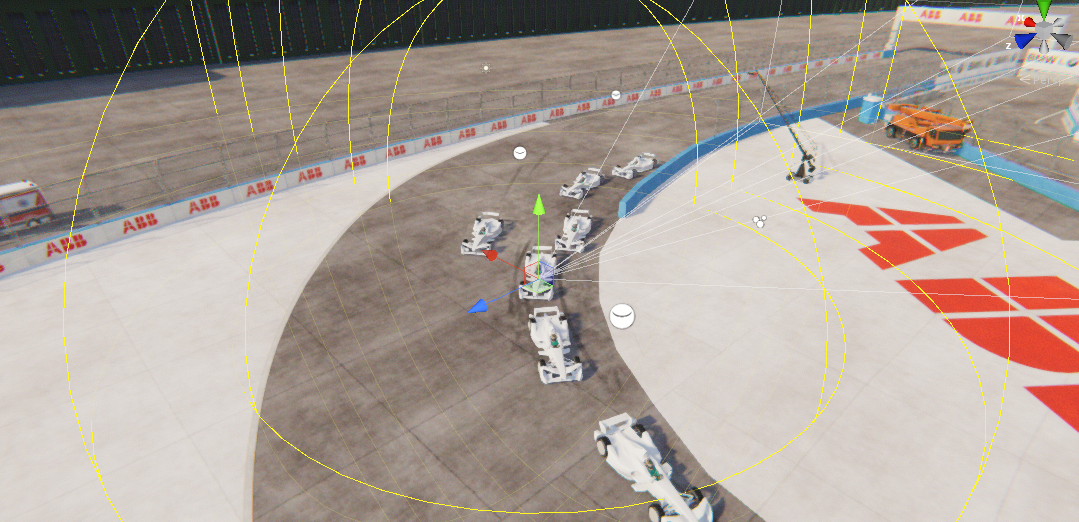} 
\end{center}
    \caption{Representation of \textit{Dome} mode setup with a set of cars placed with a predefined arrangement in a fictitious environment. The yellow lines represent the camera path to take shots, according to the configuration file.}
\label{fig:dome_placed}
\end{figure}

We developed a tool named \textit{SynGen}, using the Unity game engine \cite{shafaei2016play}, with two modes of operation:

The \textit{Dome} mode, where a static scene is recreated in the middle of a 3D modeled environment and the camera points toward a predefined object with the possibility of changing different parameters from one shot to another, such as object of interest, textures, lighting and camera parameters. This way, we have an updated and varied dataset with minimal effort. 

The \textit{Sequence} mode, containing a simple logic that roughly mimics the object’s behavior in reality inside a finite environment (usually a bit larger than the \textit{Dome}'s). It has different options to set up the simulation, such as the time to run the scene, the frame rate to capture shots and different parameters to handle the new logic for the cameras.

\textit{SynGen} is designed to be executed via JSON configuration files. For re-usability purposes these are divided into five types:

\begin{itemize}
\itemsep0em 
    \item \textbf{Output}, to define path, formats and information to store as a result of the generation.
    \item \textbf{Bundles}, to link to the graphic assets to be used during the generation.
    \item \textbf{Dome}, to specify a particular configuration of a \textit{Dome} environment. 
    \item \textbf{Sequence}, to specify all the parameters for the execution of one \textit{Sequence}.
    \item \textbf{Profiles}, a list of \textit{Domes} and \textit{Sequences} and a set of general parameters to define a dataset generation run.
\end{itemize}

Once the batch of configuration files is prepared, generating a new dataset can ony take a few hours depending on the hardware. The operation can also be run in parallel by clustering hardware resources, reducing the time to just a few minutes.

\subsection{Iteration 1. Information relative to the real environment}

\begin{figure}[t]\begin{center}
\includegraphics[width=1\linewidth]{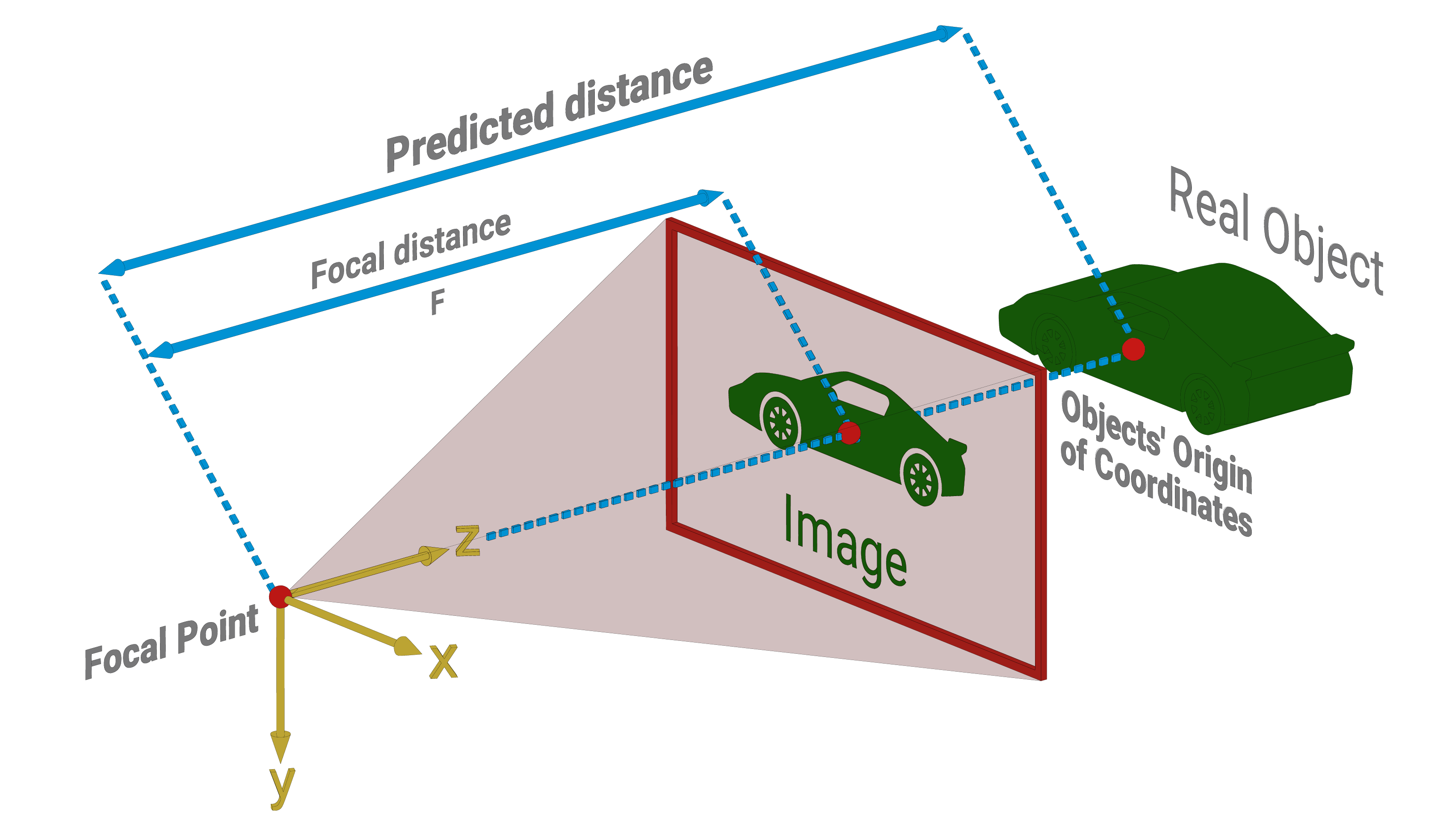} 
\end{center}
\caption{Obtaining object's position in the real world}
\label{fig:focal}
\end{figure}

This first iteration of the system (\textit{Iteration1}) is able to predict the 3D position of objects w.r.t. the camera’s default coordinate system. That way it provides valuable information in any context where we need 3D spatial awareness, such as scene recognition or SLAM (Simultaneous Localization and Mapping) \cite{handa2014benchmark}.

The modifications to the NN used in the original system consist of adding (a) the position in the image of the object’s origin of coordinates. By design, the object is inscribed in a cuboid and the origin of coordinates is at the center of the cuboid’s lower face; and (b) the distance from the camera focus on the object’s origin of coordinates.

In Deep Learning applied to object detection the use of anchor boxes (or prior boxes) is a prevalent strategy and many NN architectures use them as the basis for their predictions. These priors consist of a series of \textit{k} default bounding boxes that are used on each cell of the predicted feature map as templates to decode the actual object’s coordinates on the frame. Thus, for each cell, the NN proposes \textit{k} object candidates, whose locations are given as a shift of the current cell position and the size of the predefined priors. The reasoning behind this is based on the idea that it is easier for a NN to learn small values or variations over a predefined item than quantities in a bigger range \cite{Zhong_2020_WACV}. In practice, the priors act as a reference or prototype of format factor and size of the objects that are desired to be detected.

As this system is largely based on YOLOv3 \cite{yolov3}, the priors are necessarily the same, with some additions to serve our purpose. For each anchor (having default values for bounding box width and height), we need anchor values for (a) and (b) described before and activation functions. YOLO's bbox predictions (parameters for center and size of bbox, $t_x,t_y,t_w,t_h$) use sigmoids and exponentials. In this sense, (b) uses exponential, just like width/height (like $b_w=p_w \cdot e^{t_w}$). (a), while in principle might benefit from the same approach, was found somewhat unstable, so a sigmoid with a suitably adjusted range was used instead (similar to $b_x = c_x + \sigma_0(t_x)$ where $\sigma_0(\cdot)$ is a modified version of the sigmoid).

The two additional essential details that we need from \textit{SynGen}, compared to the original system, are:

\begin{enumerate}
    \item The Field of View (FOV) or scale of the camera system.
    \item For each object, the position in the image of its origin of coordinates, as well as the distance from the camera focus to said origin of coordinates.
\end{enumerate}

\subsection{Iteration 2. Orientation}

In the second iteration (\textit{Iteration2}), the system provides the same inferred information while also giving the orientation of the detected objects. This implies a more robust realization of 3D spatial awareness. For example, it adds to a more robust object tracking and movement prediction, not only in the camera plane, but spatially around the camera. Additionally, we can focus our attention to analyze specific parts of the detected object.

This new method represents a radical departure from the original architecture. The detection network remains the same, but the prior system and, to a great extent, the loss function is completely redesigned. Once we have the object’s orientation, distance and position of its origin on the image, we can interpret any point in the object’s reference frame and re-project it to the camera plane (or vice-versa). This enables us to overlay 3D and 2D bounding boxes on the objects.

\subsubsection{Training data}

\begin{figure}[t]\begin{center}
\includegraphics[width=1\linewidth]{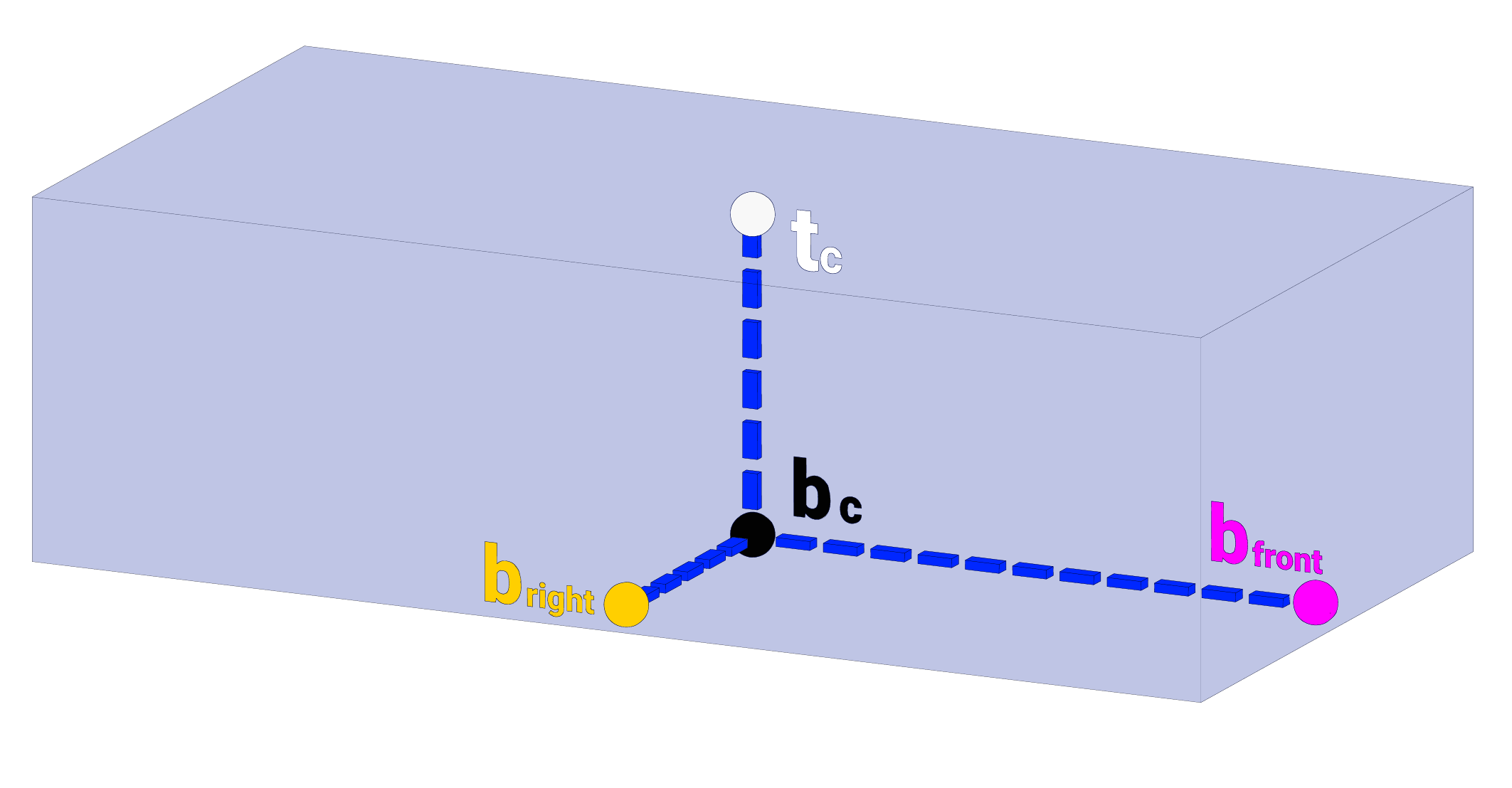} 
\end{center}
\caption{3D bounding box definition}
\label{fig:cuboid}
\end{figure}

3D information should be in the training data. Additionally, all the images should have the same FOV and aspect ratio or, at the very least, the same aspect ratio and very similar FOVs within a tight tolerance. In the current iteration, all objects should have the same actual size, but allowing for varying sizes would be a matter of also predicting the scale (in each dimension, if necessary). In this iteration, we require the same information as in the previous step, plus the orientation of each object. Both the distance and the orientation can be computed if we have the 3D coordinates (w.r.t. the camera's coordinate system) of four identified 3D points of the object. In practical terms, we have settled on the following 3D points of the 3D bounding box:

\begin{itemize}
\itemsep0em 
    \item[--] Center of the base of the cuboid (this will mark our origin of coordinates).
    \item[--] Center of the right side of the base of the cuboid (this will mark the object's X direction).
    \item[--] Center of the front side of the base of the cuboid (this will mark the object's Y direction).
    \item[--] Center of the top of the cuboid (this will mark the object's Z direction).
\end{itemize}

We added three more fields in the annotation files produced by \textit{SynGen}: one to describe the cuboid enclosing the object, using the camera as the origin of its coordinate system, another one using the 3D environment center as the origin and a third one with the object’s cuboid points projected onto the camera screen to avoid some computation at training time (see Figure~\ref{fig:cuboid}).

\subsubsection{Priors}

In \textit{Iteration1}, we achieved quick convergence on the training with high accuracy on the use-case application. For this reason, we decided to build the next step on the same principles. Samples are not grouped by bounding box now. Instead, they are first grouped by the neural network head (coarser heads are used for nearer objects), then by orientation/distance and finally, by each anchor. The network predicts the same information as before (position in the image of the origin of coordinates, distance to the camera), as well as the orientation of the object, but not explicit data about the bounding boxes.

\begin{figure}[hbt!]
\centering
    \begin{subfigure}{.46\textwidth}
        \centering
        \includegraphics[width=.8\linewidth]{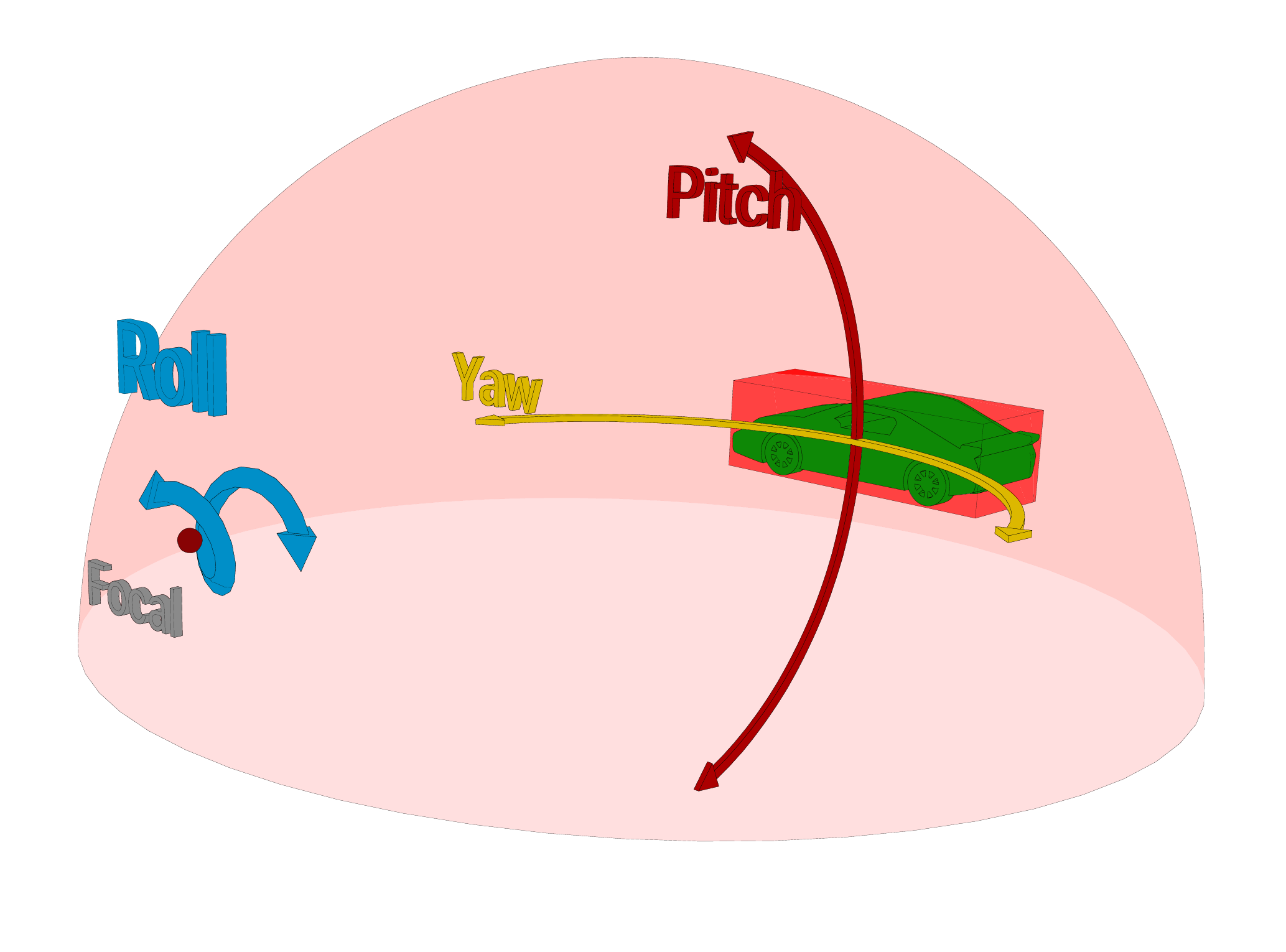}
        \caption{Angular components of the object’s cuboid hull}
        \label{fig:euler-angles}
    \end{subfigure}
    \begin{subfigure}{.46\textwidth}
        \centering
        \includegraphics[width=.8\linewidth]{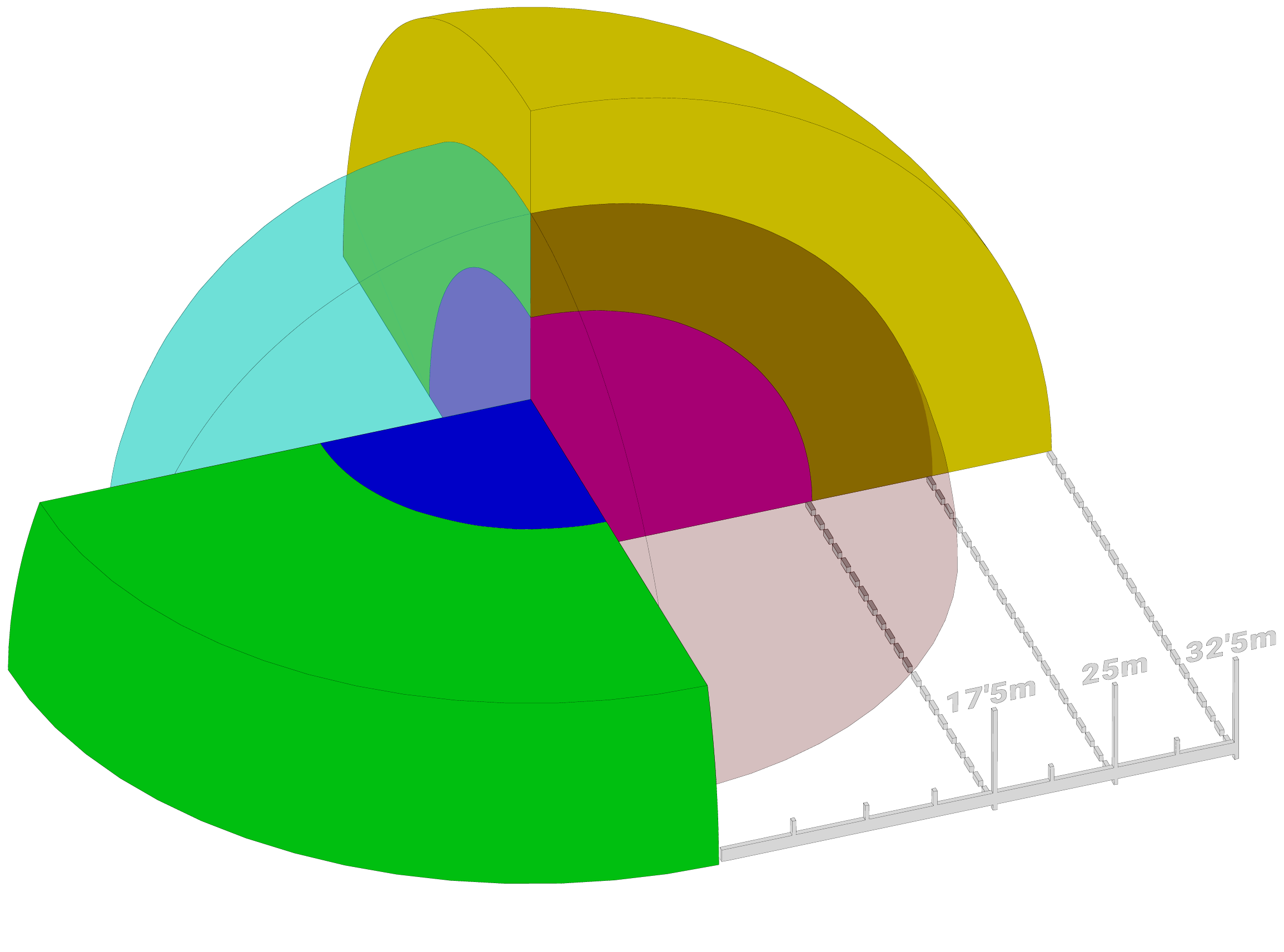}
        \caption{Regions assigned to the 6 nearest priors}
        \label{fig:angles-priors}
    \end{subfigure}
    \caption{Drawing (a) shows a representation of the angular components, based on extrinsic Euler angles, to describe the object's cuboid hull to be inferred by the neural network. Drawing (b) represents the regions assigned to the 6 priors for head 1. The two front-facing regions (blue and green) are horizontally sectioned for clarity.}
    \label{fig:angles}
\end{figure}

We defined 18 priors, using a semi-automated procedure. We set up and ran multiple simulations using \textit{SynGen} and generated a large enough set of samples, representative for the use-case that we intended to tackle. Then, studied groupings of objects’ locations (defined by their distance and rotational parameters w.r.t the camera that took the shot on each sample). Then, balancing different criteria, we divided the space into 6 regions per head, 18 in total. First distributed into the three neural network heads, then four regions per head according to the yaw component (front, back, left, and right of the object) and dividing once again by distance the front and back sides (see Equation~\ref{eq:priors} and Figure~\ref{fig:angles} for a visual representation). Angle components yaw, pitch and roll of the anchors were established by computing the average \cite{so3} of the samples falling in each region.

With the sphere equation, the regions for head 1 were defined as follow ($r$ corresponds to the distance between the object and the camera, $\theta$ to the yaw and $\phi$ to the pitch):

$x = r \cdot cos\phi \cdot cos\theta,\ y = r \cdot cos\phi \cdot sin\theta,\ z = r \cdot sin\phi$

\begin{equation}
    \begin{array}{lll}
    0 \leq r < 17.5, & -\pi/8 \leq \theta < \pi/8, & 0 \leq \phi\\
    17.5 \leq r < 32.5, & -\pi/8 \leq \theta < \pi/8, & 0 \leq \phi \\
    0 \leq r < 25, & \pi/8 \leq \theta < 3\pi/8, & 0 \leq \phi \\
    0 \leq r < 17.5, & 3\pi/8 \leq \theta < 5\pi/8, & 0 \leq \phi \\
    17.5 \leq r < 32.5, & 3\pi/8 \leq \theta < 5\pi/8, & 0 \leq \phi \\
    0 \leq r < 25, & 5\pi/8 \leq \theta < 7\pi/8, & 0 \leq \phi \\
    \end{array}
    \label{eq:priors}
\end{equation}

Then the regions for head 2 were defined similarly, with delimiting radius 32.5, 50, 70, for front and back and 25, 60 for the lateral regions. The same for head 3, with 70, 90, 110 and 60, 100, respectively.

\subsubsection{Training}

Typical Deep Learning systems are geared towards loss functions implicitly using Euclidean metrics, even for no other reason than easiness and familiarity on the implementer's side. Given that a large aspect of each iteration's anchor system is predicting shifts from a default orientation, it is worth paying attention to this part of the problem lest we end up forcing our network to learn SO(3)'s uncanny topology at the probable expense of significantly diminished prediction power. SO(3) can be approximated as Euclidean for sufficiently small regions. We tile the anchors so that each one covers a solid angle as small as practically possible. We decided to use extrinsic Euler angles to encode orientations, as these translate small changes in object orientation to small changes in object appearance in a straightforward and intuitive way. The network's input is the object's appearance, after all. It's worth noting that with Euler angles, we have to be very careful to avoid gimbal lock issues when setting up rotation anchors \cite{brezov2013new}.

Because of the inherent instability of training systems with exponential activation functions and due to the willingness to forgo detection of objects absurdly close/far to the camera, we decided to use sigmoid activation \cite{karlik2011performance} with ranges suited to each specific requirement. If detection of very close/far objects is required, we can and should use exponential activation for distance predictions, just like the original neural network did for the bounding box's width and height. The loss function also had to be tailored w.r.t. classic YOLOv3. Specifically, objectness loss has two components (a) Reinforce detection of samples in the grid position where they are actually located and (b) penalize detections whose position, size and orientation are too far off from any of the actual samples present in the image. One mechanism to decide whether the orientation is too far, is to require the detecting anchor to be the same as the anchor in the ground truth. But also, in classic YOLOv3 to decide if the position and size is too far off from the true sample, they use a threshold on the maximum of the IoUs (Intersection-over-Union) between the predicted bounding box and all the bounding boxes of the true samples present in the image \cite{rezatofighi2019generalized}. While we may think of a similar scheme but using IoUs of 3D boxes, there are a lot of complicating factors:

\begin{itemize}
\itemsep0em 
    \item[--] By design, the boxes are 3D and they are not axis-aligned. This makes 3D IoU computations much harder than 2D IoUs. \cite{zhou2019iou}
    \item[--] It weakens the signal when the predicted distance is too far off the actual distance, as the 3D IOU can get down to zero pretty fast, while the actual effect in the image of a different distance is a (possibly minimal) difference in object size.
    \item[--] Object's image size is inversely proportional to the distance to the camera, so the relation between the network's input (the image) and the predicted variables is unnecessarily more non-linear. This translates as "harder to predict".
\end{itemize}

Due to these issues, we decided to continue with the same approach as in the original neural network. Since only 3D data is predicted, we need to compute a 2D version of the bounding boxes:

\begin{itemize}
\itemsep0em 
    \item[--] For each object, we define a set of points in its internal reference frame, representing its convex hull or a good enough approximation to it.
    \item[--] For each predicted object, with its position in the screen, distance and orientation, we compute the 3D positions (in the camera's reference frame) of each of these points.
    \item[--] We re-project these 3D points into the camera's plane. Taking the max/min in both axes, we get quite a good approximation to the object's bounding box.
\end{itemize}

\subsection{Iteration 3. Information aggregation and self-calibration}

\begin{figure}[t]\begin{center}
\includegraphics[width=1\linewidth]{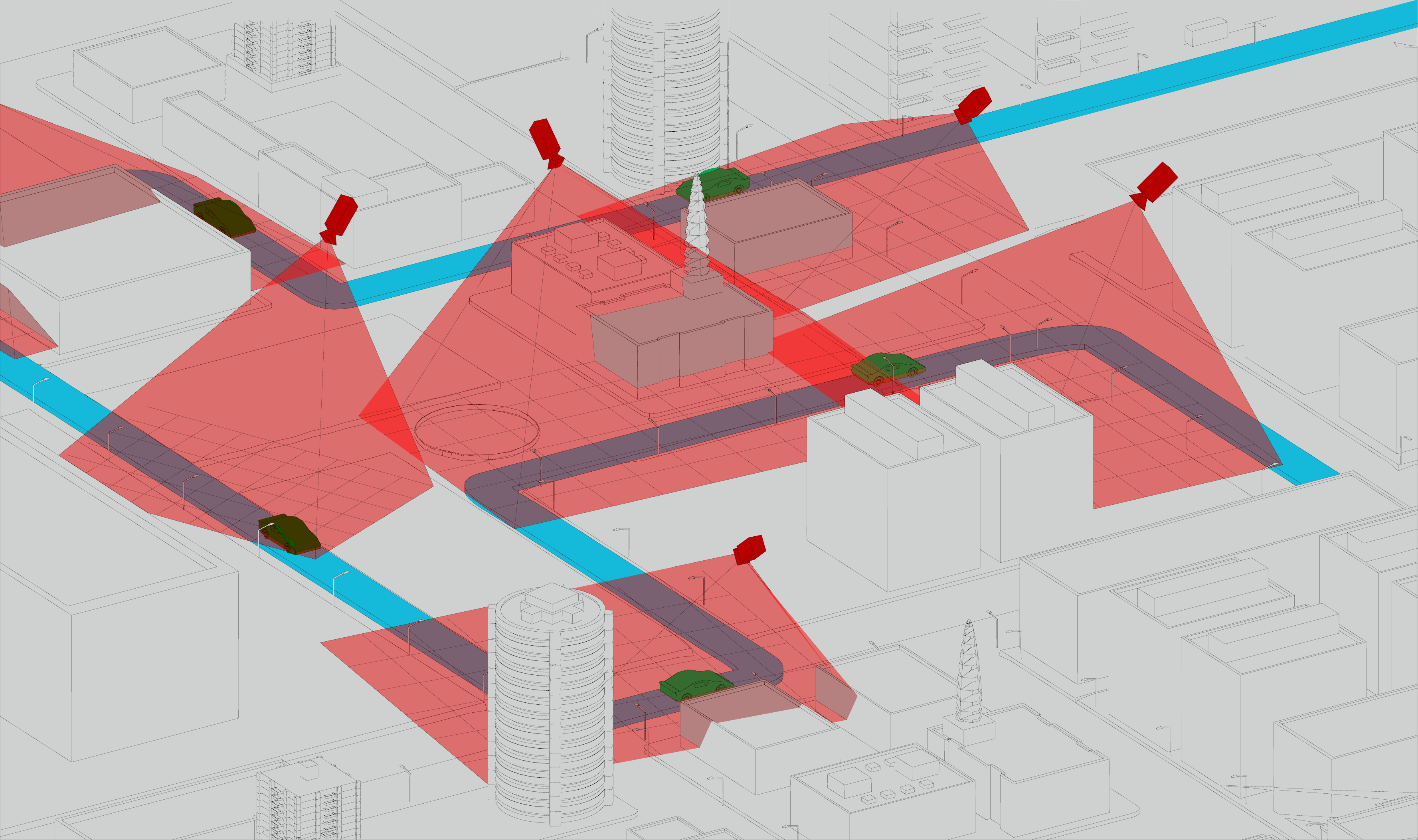} 
\end{center}
\caption{Circuit covered by multiple cameras with overlapping regions}
\label{fig:aggregation}
\end{figure}
This approach is intended to increase the accuracy of the detected locations and minimize losing track, by overlapping the areas covered by the different cameras. At the same time, a system is being designed which will be able to infer useful information about the environment, such as the plane of movement of the objects. It will increase the accuracy and ultimately enable self-calibration. The system will not have to rely on the cameras' GPS to compute actual positions, it will instead rely on the detection of descriptive points, SLAM running in coordination within the system. The system is currently working under a simple heuristic: the predicted location to provide for each object is the one whose inferred distance to the camera is the lowest, which tends to be the most accurate one. Here the precision of the system is strictly determined by the accuracy of the detecting method built in phase 2.
        
\section{Experiments}
\label{section_experiments}

\subsection{Training setup}

The software setup consists of Kubuntu 18.04, Python, Keras, Tensorflow, CUDA and other miscellaneous software required. We train using single GPU accross different machines with a similar configuration: CPU Intel i9-9820X, memory 64GB, 2 x GPU RTX 2080Ti. We performed a batch of experiments on both \textit{Iteration1} and \textit{Iteration2} where we handled the different considerations as follows:

\begin{itemize}
\itemsep0em 

    \item The same pre-trained weights are always loaded, since it accelerates the training and there is not much difference in convergence w.r.t. loading pre-trained weights from any other of our experiments and it sets a clear and common baseline to all of them.
    \item The generated datasets are always distributed in the same fashion, 80\% to the training set, 10\% to validation and 10\% to test.
    \item The priors in \textit{Iteration2} are independent of image size and once they were computed we used the same for all the experiments.
    \item The batch size is set up to 8, since it provides good convergence and the training can be done using one of the described GPU.
    \item The metric to inspect and control the training process was the validation loss (all its components were displayed separately on TensorBoard, which was useful to debug loss functions and datasets).
    \item The learning rate is set up with an exponential decay function of fixed parameters ($lr = 10^{-3} / (1 + epoch^{1.5})$).
    \item Most of the hyper-parameters were set up to their final values one by one and then we focused our experiments mainly on the improves over the datasets: variety and quality of the models and the virtual environments. We also performed tests with images at two different resolutions (see Table~\ref{table:datasets}), consequently scaling up the network.
    \item \textit{SynGen} already provides highly varied sets of samples for training, making less relevant the use of standard image augmentation techniques. The only two operations applied in our tests were changes of exposure and saturation. Actually, other common alterations, such as change of scale or aspect ratio would be impractical due to the nature of our use case.
\end{itemize}

\begin{table}
    \caption{Datasets used for training. It shows total number of samples, frame resolution in pixels used at the input of the neural network, percentage of images generated with the environment \textit{Dome} and percentage with \textit{Sequence}.
    }
    \label{table:datasets}
    \centering
    \resizebox{\columnwidth}{!}{
    \begin{tabular}{lrrrr}
        \toprule
        id & \#samples & resolution & \%Dome & \%Seq \\
        \midrule
        syn-s05-h81-se & 36741 & 416x416 & 0 & 100 \\
        syn-s05-h91-se & 370466 & 416x416 & 0 & 100 \\
        syn-s05-h92-se & 370466 & 832x832 & 0 & 100 \\
        syn-s05-h93-se & 370466 & 1472x832 & 0 & 100 \\
        syn-s06-h101-ds & 407626 & 736x416 & 37 & 63 \\
        syn-s06-h102-ds & 407626 & 1472x832 & 37 & 63 \\
        \bottomrule
    \end{tabular}
    }
\end{table}

\subsection{Training results}

We trained both iterations of our system with synthetic images only (see Table~\ref{table:datasets}). In the first experiments with \textit{Iteration1}, using \textbf{syn-s05-h81-se} and \textbf{syn-s05-h91-se}, we quickly found out that the first dataset was too small, metrics with the test set were worse and the convergence indicated the same. While in the first set of tests we only achieved a validation loss of 7.41 in the best experiment after more than 20 epochs, in the second case, the same metric achieved stable values between 2 and 3, usually well before 20 epochs. 
A last batch of experiments were done with \textbf{syn-s05-h92-se}, having the same images but scaling up the input to 832x832. With regard to the training times, while with small images \textit{Iteration1} would spend around 37 minutes per epoch, per 100k samples in the dataset, using the hardware specified above, it would be in the range of 427 minutes in the case of large images. Stabilizing in similar values of the reference metric would take longer as well, more than 50 epochs and, although it is true that the output was of a better quality, we never achieved the expected accuracy; \textit{Iteration1} predicted the position of the object's origin of coordinates quite faithfully, but distance prediction for medium-large distances was relatively poor (slightly above 1 meter in mean error for objects at distances between 50 and 60 meters). We think that this might have been an artifact of changing the image's aspect ratio, as this introduces an additional and quite pronounced trigonometric non-linearity on the relationship between an object's apparent size and distance.

With \textit{Iteration2}, training time is around 45 minutes per epoch, per 100k samples using small images (\textbf{syn-s06-h101-ds}) and 530 minutes using the large version (\textbf{syn-s05-h93-se}, \textbf{syn-s06-h102-ds}). It takes a little bit longer than with \textit{Iteration1}, essentially due to the difference on the image size. \textit{Iteration2} was conceived to preserve the image’s aspect ratio, hence at small sizes for input images, the orientation predictions were already good and the distance predictions were very good: in the order of tens of centimeters for long distances. However, with the change of resolution the increase in the accuracy was very noticeable, validation loss would stabilize in much lower values (up to 3.95, as opposed to 7.62, the minimum achieved with small images) and distance prediction error for medium distance, between 50 and 60 meter, was in the order of centimeters.


\begin{table}
    \caption{Per-frame inference times (ms) for small (416x736) and large (832x1472) images}
    \label{table:inferencetimes}
    \centering
    \begin{tabular}{lcccc}
        \toprule
                    & \multicolumn{2}{c} {Dev-Machine}    & \multicolumn{2}{c} {Production-Machine}    \\
        \cmidrule(r){2-3}
        \cmidrule(r){4-5}
                    & small         & large             & small         & large             \\
        \midrule
        TensorFlow  & 50            & 160               & 26            & 85                \\
        Tensor RT   & 25            & 80                & 12            & 42                \\
        \bottomrule
    \end{tabular}
\end{table}

We have implemented the inference in a pipeline of five processes: capture, pre-processing, inference, post-processing and serving. For a same image size, time spent on the GPU is almost identical for \textit{Iteration1}, \textit{Iteration2} and the original neural network, as the only changes at GPU level are different sizes in the output tensors. An optimized C++ pre-processing and a vectorized and optimized post-processing make the inference the heaviest process of all by far, hence determining the final throughput of the system. Table~\ref{table:inferencetimes} shows the results of tests performed on the development (Intel i7-7700K, 32Gb RAM and a GTX 1080) and production machines (Intel i9-9820X, 128Gb RAM and a RTX 2080Ti).

\section{Conclusion}
\label{section_conclusion}

We have presented the development of a fully operational application that employs forefront AI as a service. We have managed to built a positional tracking system of high precision, where 25cm is the worst case in production, and high update rate, currently around 24Hz with regular hardware, which can be increased just with more computing power, while conventional tracking systems work at the scale of 10Hz and are limited by the technology. We sorted out the problem of training a system like this by using only synthetic imagery, achieving the desired performance on real cases. By design, the detecting precision is scalable, as experiments shows, and resolution is one way of doing it; another way is by using more cameras covering the area of interest, thus any point will be closer to a camera; and third, by interchanging the neural network backbone with a newer architecture, such as EfficientDet \cite{tan2019efficientdet} or YOLOv4 \cite{bochkovskiy2020yolov4}. Lastly, it is worth mentioning that there is an actual very promising development consisting of moving from the simple heuristic to handle information inferred from multiple cameras, to a NN-based aggregation approach, which would increase accuracy and overall performance.

Since the system is essentially based on AI executing computer vision over video footage, the hardware is fairly standard and inexpensive, this is true as well for the cameras. Their placement is also very flexible and can work outdoor and, more importantly, indoor, where the GPS cannot operate. Although our current application is sport-related, the versatility of this system and the possibility to offer it both as a local setup or through a cloud-based service, make the range of uses huge. In particular, for indoor/outdoor positioning of personnel and objects in factories, shopping malls, rail-stations, hospitals, museums, etc. with many diverse targets, such as personal assistance, smart-apps, security, increasing efficiency of processes, etc.

{\small
\bibliographystyle{ieee}
\bibliography{references}

\begin{thebibliography}{10}\itemsep=-1pt

\bibitem{bochkovskiy2020yolov4}
A.~Bochkovskiy, C.-Y. Wang, and H.-Y.~M. Liao.
\newblock Yolov4: Optimal speed and accuracy of object detection.
\newblock {\em arXiv preprint arXiv:2004.10934}, 2020.

\bibitem{brezov2013new}
D.~S. Brezov, C.~D. Mladenova, and I.~M. Mladenov.
\newblock New perspective on the gimbal lock problem.
\newblock In {\em AIP Conference Proceedings}, volume 1570, pages 367--374.
  American Institute of Physics, 2013.

\bibitem{dubuisson2016survey}
S.~Dubuisson and C.~Gonzales.
\newblock A survey of datasets for visual tracking.
\newblock {\em Machine Vision and Applications}, 27(1):23--52, 2016.

\bibitem{handa2014benchmark}
A.~Handa, T.~Whelan, J.~McDonald, and A.~J. Davison.
\newblock A benchmark for rgb-d visual odometry, 3d reconstruction and slam.
\newblock In {\em 2014 IEEE international conference on Robotics and automation
  (ICRA)}, pages 1524--1531. IEEE, 2014.

\bibitem{so3}
R.~Hartley, J.~Trumpf, Y.~Dai, and H.~Li.
\newblock Rotation averaging.
\newblock {\em International journal of computer vision}, 103(3):267--305,
  2013.

\bibitem{karlik2011performance}
B.~Karlik and A.~V. Olgac.
\newblock Performance analysis of various activation functions in generalized
  mlp architectures of neural networks.
\newblock {\em International Journal of Artificial Intelligence and Expert
  Systems}, 1(4):111--122, 2011.

\bibitem{koyuncu2010survey}
H.~Koyuncu and S.~H. Yang.
\newblock A survey of indoor positioning and object locating systems.
\newblock {\em IJCSNS International Journal of Computer Science and Network
  Security}, 10(5):121--128, 2010.

\bibitem{langley1998rtk}
R.~B. Langley.
\newblock Rtk gps.
\newblock {\em GPS World}, 9(9):70--76, 1998.

\bibitem{contextualnet}
M.~Patel, B.~Emery, and Y.-Y. Chen.
\newblock Contextualnet: Exploiting contextual information using lstms to
  improve image-based localization.
\newblock In {\em 2018 IEEE International Conference on Robotics and Automation
  (ICRA)}, pages 1--7. IEEE, 2018.

\bibitem{yolov3}
J.~Redmon and A.~Farhadi.
\newblock Yolov3: An incremental improvement.
\newblock {\em arXiv preprint arXiv:1804.02767}, 2018.

\bibitem{rezatofighi2019generalized}
H.~Rezatofighi, N.~Tsoi, J.~Gwak, A.~Sadeghian, I.~Reid, and S.~Savarese.
\newblock Generalized intersection over union: A metric and a loss for bounding
  box regression.
\newblock In {\em Proceedings of the IEEE Conference on Computer Vision and
  Pattern Recognition}, pages 658--666, 2019.

\bibitem{salih2013suitability}
A.~A. A.-A. Salih, N.~L. A. C.~A. Zaini, and A.~Zhahir.
\newblock The suitability of gps receivers update rates for navigation
  applications.
\newblock In {\em Proceedings of World Academy of Science, Engineering and
  Technology}, page 192. World Academy of Science, Engineering and Technology
  (WASET), 2013.

\bibitem{shafaei2016play}
A.~Shafaei, J.~J. Little, and M.~Schmidt.
\newblock Play and learn: Using video games to train computer vision models.
\newblock {\em arXiv preprint arXiv:1608.01745}, 2016.

\bibitem{scenecoord}
J.~Shotton, B.~Glocker, C.~Zach, S.~Izadi, A.~Criminisi, and A.~Fitzgibbon.
\newblock Scene coordinate regression forests for camera relocalization in
  rgb-d images.
\newblock In {\em Proceedings of the IEEE Conference on Computer Vision and
  Pattern Recognition}, pages 2930--2937, 2013.

\bibitem{smeulders2013visual}
A.~W. Smeulders, D.~M. Chu, R.~Cucchiara, S.~Calderara, A.~Dehghan, and
  M.~Shah.
\newblock Visual tracking: An experimental survey.
\newblock {\em IEEE transactions on pattern analysis and machine intelligence},
  36(7):1442--1468, 2013.

\bibitem{sun2015generating}
B.~Sun, X.~Peng, and K.~Saenko.
\newblock Generating large scale image datasets from 3d cad models.
\newblock In {\em CVPR 2015 Workshop on the future of datasets in vision},
  2015.

\bibitem{tan2019efficientdet}
M.~Tan, R.~Pang, and Q.~V. Le.
\newblock Efficientdet: Scalable and efficient object detection.
\newblock {\em arXiv preprint arXiv:1911.09070}, 2019.

\bibitem{yilmaz2006object}
A.~Yilmaz, O.~Javed, and M.~Shah.
\newblock Object tracking: A survey.
\newblock {\em Acm computing surveys (CSUR)}, 38(4):13--es, 2006.

\bibitem{Zhong_2020_WACV}
Y.~Zhong, J.~Wang, J.~Peng, and L.~Zhang.
\newblock Anchor box optimization for object detection.
\newblock In {\em The IEEE Winter Conference on Applications of Computer Vision
  (WACV)}, March 2020.

\bibitem{zhou2019iou}
D.~Zhou, J.~Fang, X.~Song, C.~Guan, J.~Yin, Y.~Dai, and R.~Yang.
\newblock Iou loss for 2d/3d object detection.
\newblock In {\em 2019 International Conference on 3D Vision (3DV)}, pages
  85--94. IEEE, 2019.

\end{thebibliography}
}

\end{document}